\documentclass{wiley-article}

\usepackage[numbers,super]{natbib}
\usepackage{siunitx}

\usepackage[toc,page]{appendix}
\usepackage{subfigure}
\usepackage{amsmath}
\usepackage{graphicx}
\usepackage{multirow}
\usepackage{amssymb}

\papertype{Research Article}
\paperfield{}

\title{A Data-driven Adversarial Examples Recognition Framework via Adversarial Feature Genomes}

\author[1]{Li~Chen}
\author[2]{Qi~Li}
\author[3]{Weiye~Chen}
\author[4]{Zeyu~Wang}
\author[1]{Haifeng~Li}

\affil[1]{School of Geosciences and Info-Physics, Central South University, South Lushan Road, Changsha, 410083, China.}
\affil[2]{School of Computer Science and Engineering, Central South University, South Lushan Road, Changsha, 410083, China.}
\affil[3]{Department of Geography and Geographic Information Science, University of Illinois at Urbana-Champaign, USA.; CyberGIS Center for Advanced Digital and Spatial Studies, University of Illinois at Urbana-Champaign, USA.}
\affil[4]{School of Civil Engineering, Central South University, South Lushan Road, Changsha, 410083, China.}

\corraddress{Haifeng Li}
\corremail{lihaifeng@csu.edu.cn}

\fundinginfo{This work was supported by the National Natural Science Foundation of China (Grant Numbers 41871364, 41871276, 41871302, and 41861048). The authors also appreciate the High Performance Computing Platform of Central South University and HPC Central of Department of GIS to provide HPC resources. \\ Citation: \par
 Li Chen, Qi Li, Weiye Chen, ZeyuWang, Haifeng Li. A Data-driven Adversarial Examples Recognition Framework via Adversarial Feature Genomes. International Journal of Intelligent Systems. 2022. doi:10.1002/int.22850.}

\runningauthor{Li Chen, et al.}

\begin{document}

\begin{frontmatter}
\maketitle

\begin{abstract}
Adversarial examples pose many security threats to convolutional neural networks (CNNs). Most defense algorithms prevent these threats by finding differences between the original images and adversarial examples. However, the found differences do not contain features about the classes, so these defense algorithms can only detect adversarial examples without recovering the correct labels. In this regard, we propose the Adversarial Feature Genome (AFG), a novel type of data that contains both the differences and features about classes. This method is inspired by an observed phenomenon, namely the Adversarial Feature Separability (AFS), where the difference between the feature maps of the original images and adversarial examples becomes larger with deeper layers. On top of that, we further develop an adversarial example recognition framework that detects adversarial examples and can recover the correct labels. In the experiments, the detection and classification of adversarial examples by AFGs has an accuracy of more than 90.01\% in various attack scenarios. To the best of our knowledge, our method is the first method that focuses on both attack detecting and recovering. AFG gives a new data-driven perspective to improve the robustness of CNNs. The source code is available at https://github.com/GeoX-Lab/Adv\_Fea\_Genome.

\end{abstract}
\end{frontmatter}

\section{Introduction}

Convolutional neural networks (CNNs) have achieved remarkable success in a variety of tasks~\cite{lecun2015deep}. It has become a fundamental component of many computer vision tasks. However, most CNNs are suscept to adversarial examples~\cite{SzegedyZSBEGF13,9339955,int.22260} which leads to high-confidence misclassification, resulting from small crafted perturbation to the original image. The effect of adversarial examples testifies that CNNs have serious security issues despite their excellent performance.

In order to defend against adversarial examples, two main strategies of defense algorithms are proposed and evolved, namely, complete defense and detection only~\cite{akhtar2018threat}. The complete defense aims to improve the robustness of the model~\cite{gu2014towards,wang2016learning}, thus increasing the difficulty of generating adversarial examples.
For instance, as the most commonly used defense algorithm, adversarial training~\cite{KurakinGB17,TramerKPGBM18} puts the obtained adversarial examples as new data into the original training set and retrains the CNNs. This process makes the CNNs robust to the adversarial examples. 
Given that most adversarial examples attack based on gradients, gradient masking~\cite{tramer2017ensemble} attempts to prevent the attacker from accessing a useful gradient and prevents the generation of adversarial examples. And knowledge distillation~\cite{papernot2016distillation} transfers the knowledge of the complex model to a simple CNN and classifies adversarial examples with the simple CNN.
However, these strategies either reduce the generalizability of the model or are not effective over some adversarial examples. Complete defense methods do not explicitly detect adversarial examples; as a result, they remain vulnerable to stronger attacks~\cite{athalye2018obfuscated,ZhangWLPW20}.
For another type of defense algorithm, detection only~\cite{xu2017feature,wang2018interpret,meng2017magnet} uses the differences between adversarial examples and the original images to detect potential adversarial examples and reject their further processing. Detection of these differences includes Local Intrinsic Dimensionality (LID)~\cite{Ma0WEWSSHB18} or an additional detector~\cite{MetzenGFB17}. Yet, they cannot classify adversarial examples. In addition, rejecting all potential adversarial examples may make false-positive images to be otherwise handled as adversarial examples.
In summary, both types of defense algorithms have their unresolved limitations. They can either detect adversarial examples based on differences from the original images or can only correctly classify adversarial examples based on features about the classes.
 
Therefore, acquiring features about the classes from the differences between adversarial examples and the original images is the key to detect and correctly classify adversarial examples simultaneously. Through surveys and experiments, we observe linkage between the evolutions of the features in the CNNs and the differences of adversarial examples from original images~\cite{int.22458,YinWWTW20}. Although the original images and adversarial examples are difficult to be perceived by humans, they have exhibited increasing differences in their feature maps in the feedforward process of the model. We call it the Adversarial Feature Separability (AFS). The AFS reflects the effect of adversarial perturbations on the model, and data containing the AFS property can be used to detect adversarial examples.
For the features about the classes, we consider that the hierarchical features of the CNN have different semantic features of the classes of the input image. However, it is impractical for us to analyze all semantic features with many convolutional kernels per layer. For this problem, the group visualization method proposed by Olah et al.~\cite{olah2018building} provides a feasible way to visualize the main semantic features on each layer of CNN through group features.
We hypothesize that the group features at all layers for images and their corresponding adversarial examples could exhibit the property of the AFS. If this hypothesis holds, the group features can be used to classify adversarial examples.

Based on this hypothesis, we propose an Adversarial Feature Genome (AFG), a kind of data organized by stacking the stitched groups features of different layers. These stitched groups features in each layer like genes demonstrate the semantic features of the classes under different layers of CNN. Whether it is the original image or an adversarial example, each input image has corresponding AFG data. The AFG is a data format with multiple layers on the input images and models. Our experiments first examine the difference between the AFG of the original image and the AFG of the corresponding adversarial example and verify if it becomes larger as the layer deepens. Experimental results demonstrate that AFG has the AFS property. Then, we detect adversarial examples based on the AFGs. This detection strategy is similar to the detection-only, so we compare it with LID. The AFG-based detector performs better than other defense algorithms in a variety of attack scenarios, with an average of 9.38\% higher accuracy. The accuracies of adversarial example detection are over 90.01\% in most attack scenarios. We further verify whether the AFG has features about the classes. We use only the AFG of the original images and label them the same as the images. By training a new CNN on this AFG dataset, the classification accuracy with the new CNN decreases about 2.0\% from the original classification, which is a shred of evidence that AFG has features about the classes. When the input is the AFG of the original image, the features of all layers keep the semantic features of the same class. When the input is an adversarial example, the AFG contains the semantic features of the original class and the misclassified class. These experimental results show that the AFG has the AFS property and embodies features about classes that can be used to detect and classify adversarial examples simultaneously. 

 \begin{figure}[h]
\centering
\includegraphics[scale=0.75]{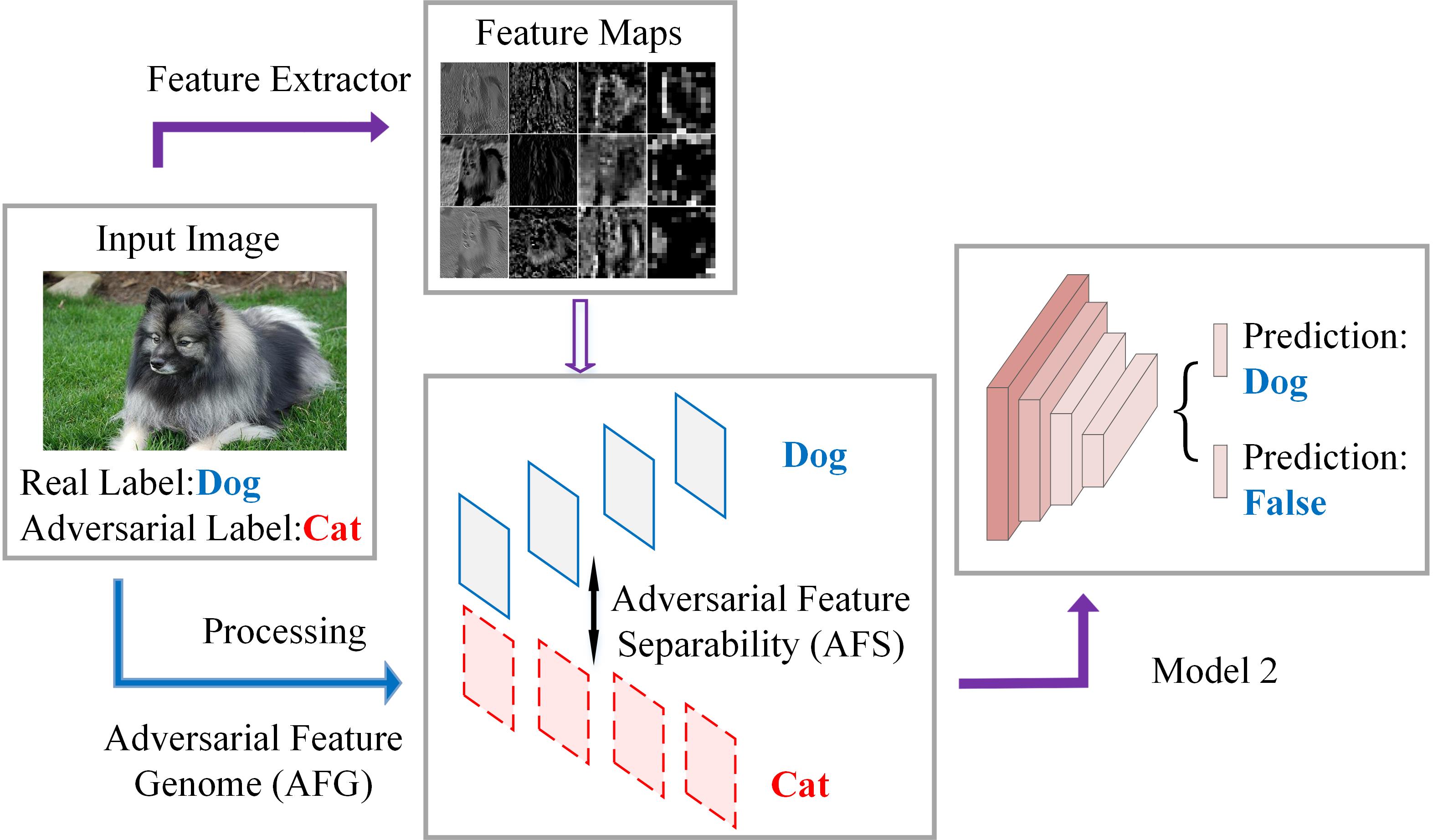}
\caption{The input image is an adversarial example, and its real label is a dog. It is misclassified as a cat by the trained CNN. In our framework, we use the trained CNN to convert the original image to an AFG. This AFG preserves the features about the classes, including the property of the AFS. We use the AFG dataset to train a second model to detect this adversarial example and classify the input image correctly.}
\label{fig:simgle}
\end{figure}

Encouraged by the findings above, we design an adversarial example recognition framework based on the AFG. It contains two modules, and the process of the framework is shown in Figure~\ref{fig:simgle}. The first one is a classification module, and the second is a defense module. An AFG also has two labels, one for the original correct class and one indicating an adversarial example. Correspondingly, the defense module is a two-path AFG recognition model. One classifier is used for the classification of the AFGs, and another is used to detect whether it is an adversarial example or not. When the framework determines that the input AFG originates from an adversarial example, its classification result is the correct label of the adversarial example. Our defense model has also demonstrated transferability for attack algorithms. Our contributions are summarized as follows:

\begin{itemize}
\item We define the AFS, a phenomenon that the difference between adversarial examples and the original image becomes larger as the layer deepens.   
\item We create AFG data, which contains features about the classes and exhibits the property of the AFS. AFG is multi-layer data, which is related to the input image and the model.
\item We propose an AFG-based adversarial examples recognition framework that can simultaneously detect and correctly classify adversarial examples. This method provides a new perspective to design defense algorithms.
\end{itemize}

\section{Related Work}
Many defense methods have been proposed to prevent the threats of adversarial examples. According to the purpose of strategy, these methods can be divided into two types: complete defense and detection only~\cite{akhtar2018threat}.

Complete defense methods aim to improve the robustness of the model by varying the input data and changing the training method. For varying the input data, the main idea is to reduce the impact of adversarial examples by destroying the structure of the adversarial perturbation. Dziugaite et al. ~\cite{dziugaite2016study} weaken the effect of adversarial perturbation by applying JPEG compression to the input images. Liao et al.~\cite{liao2018defense} generate a denoiser based on a self-encoder to filter adversarial perturbations, thus allowing defense against adversarial examples. For changing the training method, the main idea is to increase the difficulty of generating adversarial examples. For example, adversarial training~\cite{TramerKPGBM18}, the most used way for a complete defense, expands the original training set and retrains the CNNs by adding adversarial examples, thus enabling CNNs to be robust to adversarial examples. And gradient masking~\cite{tramer2017ensemble} tries to flatten the gradient of the weights to prevent the generation of adversarial examples. In addition, there is another form of attack, namely adversarial patch. These patches make the target lost or misclassified. In this regard, Chong et al.~\cite{xiang2020patchguard} design a secure feature fusion mechanism and train a CNN with a small receptive field. The model can defend against adversarial patches effectively. In summary, these complete defense methods against adversarial examples by improving the model's robustness, though at the cost of reducing the performance of CNNs. And these defenses can still be destroyed by stronger attacks~\cite{ZhangWLPW20}.

Different from obtaining a robust model, detection only methods aim to detect adversarial examples from the input images in advance~\cite{meng2017magnet,miller2020adversarial,zheng2018robust}. The essence of these methods is to defend against potential threats by detecting differences from the original image with adversarial examples and rejecting them in any further processing. They use different metrics to find the difference between the original images and adversarial examples and detect adversarial examples. Lu et al.~\cite{lu2017safetynet} and Metzen et al.~\cite{metzen2017detecting} construct different models to learn the decision boundaries of the original images and adversarial examples in the feature space to detect adversarial examples. Feinman et al.~\cite{feinman2017detecting} use uncertainty estimation of Bayesian neural networks and kernel density estimation (KDE) to detect adversarial examples. Ma et al.~\cite{Ma0WEWSSHB18} use Local Intrinsic Dimensionality (LID) to describe the dimensionality of the adversarial subspace and detect adversarial examples. Lee et al.~\cite{lee2018simple} use Gaussian discriminant analysis (GDA) to model the feature distribution and then detect adversarial examples by measuring the degree of outliers in adversarial examples based on the Mahalanobis distance (MAHA) confidence score. However, these detection-only defense algorithms do not yield the correct class about adversarial examples.

In overview, both strategies of defense algorithms have their limitations. Complete defense methods reduce the performance of CNNs, while detection only methods cannot recover the correct labels for adversarial examples. Therefore, to correctly classify adversarial examples while maintaining model performance, it is crucial to find features about the classes in the difference between the original images and adversarial examples.

\section{Adversarial Feature Separability}
Adversarial examples differ from the original images by a small perturbation. However, most CNNs produce wrong classification results when adversarial examples are input. This perturbation usually has a big impact on the processing in the feedforward of the model. Therefore, we can detect adversarial examples by identifying the difference of the input images in the feedforward process. We call the difference Adversarial Feature Separability (AFS). This difference takes many forms. For instance, Ma et al.~\cite{Ma0WEWSSHB18} find that the LID of an adversarial example is significantly higher than that of the corresponding original image. We also anticipate that the difference is shown in the feature maps during the image feedforward process.

To simplify, we use a neural network to analyze the difference between the feature maps of adversarial examples in the feedforward process and the original image.
We define $X$ as the original image and its label as $y$.
After the feedforward process, the output of the trained model generally is the same as the label of the original image. In terms of the attack, we define the generated adversarial perturbation applied to the original image as $\rho$. The original image corresponds to an adversarial example, $(X + \rho)$, for which the output of the model is not the label $y$. The process of generating adversarial example from an origin image $X$ can be defined as follows

\begin{eqnarray}\label{equ1}
\min_{\rho}\Vert\rho \Vert \quad s.t.\quad f(X+\rho)=\hat{y}.
\end{eqnarray}

Adversarial example $(X + \rho)$ makes the neural network misclassify it into class $\hat{y}$, which is different from $y$. $f(\cdot)$ represents the feedforward process of the neural network. The weights and biases of the first layer of the neural network are defined as $W$ and $b$. The difference between the feature maps of adversarial example and the original image before applying the activation function is as follows,
\begin{eqnarray}\label{equ2}
(WX+b)-(W(X+\rho)+b)=W\rho.
\end{eqnarray}

The $W\rho$ is the error caused by an adversarial example before the activation function of the first layer.
Both $W$ and $\rho$ are high-dimensional data, and Goodfellow et al.~\cite{goodfellow2014explaining} pointed out that the linearity of the high-dimensional space is responsible for adversarial examples. They argued that the cumulative errors stray from the final result due to the under-fitting of the linear portion of the deep learning model to the image. However, when the error goes through an activation function such as ReLU, its effect may be reduced because values less than 0 are filtered out. We cannot be sure whether $W\rho$ is responsible for the error in the result.
In the adversarial example problem, the predicted class $\hat{y}$ of adversarial example is not the correct label $y$. This means that the feature vectors are significantly different from that of the expected class before the classifier in a trained model. This indicates that the error $W\rho$ is passed from the first layer to the last layer.

\begin{figure}[h]
  \subfigure[]{
    \label{fig:subfig:L1} 
    \includegraphics[scale=0.21]{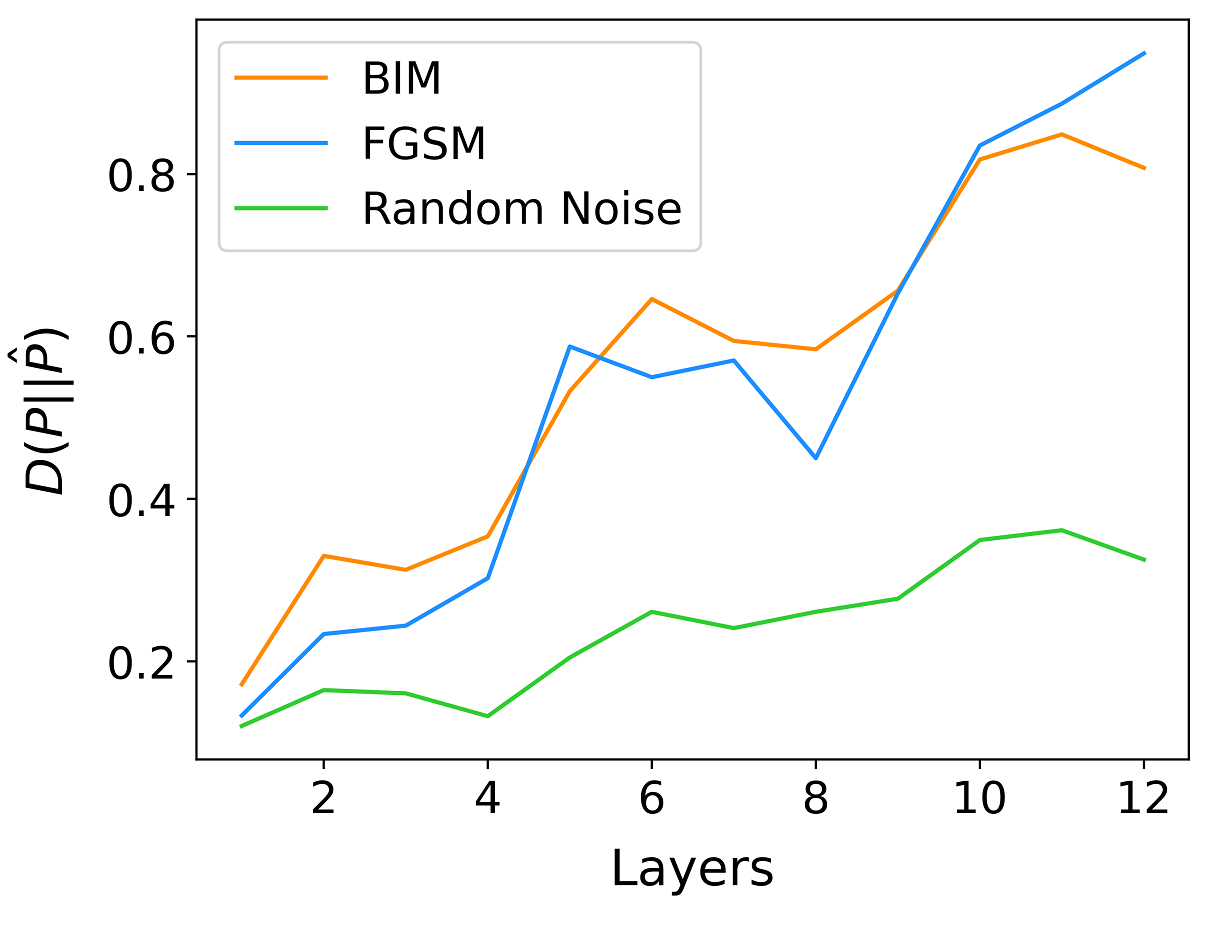}
    }
  \subfigure[]{
    \label{fig:subfig:KL} 
    \includegraphics[scale=0.18]{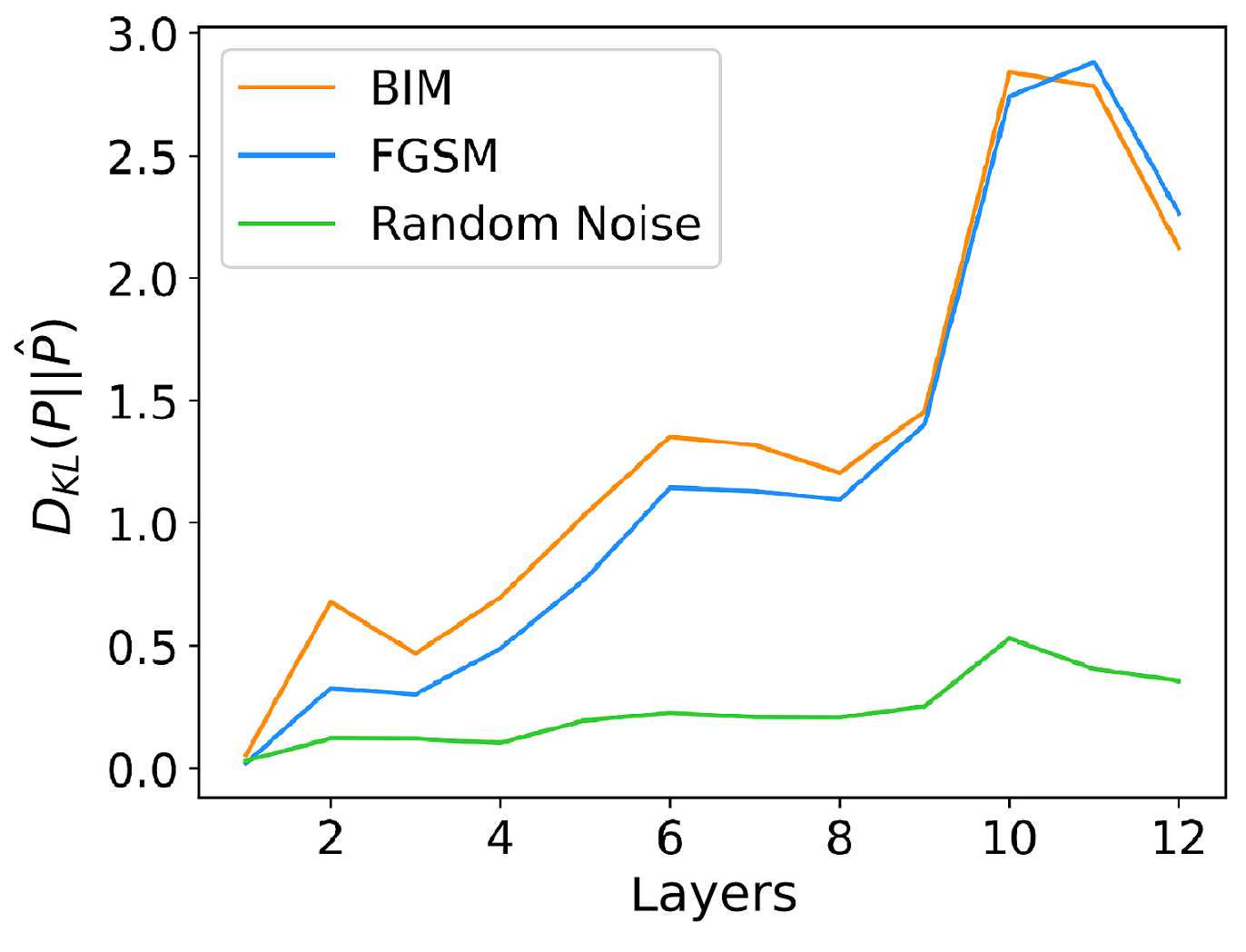}
    }
\caption{(a) The distance TSnd (b) the KL divergence to measure the difference between adversarial examples and original images on each layer with different attack algorithms. (a) and (b) show that as the layer becomes deeper, the difference in the feature maps becomes larger for all attack algorithms. The changing pattern in this difference is the AFS.}
\label{fig:distance}
\end{figure}

The feedforward process for a CNN is more complex. As in neural network analysis, the feedforward of a CNN can lead to differences in the feature maps at each layer due to adversarial perturbation.
In order to verify the difference between the original image and adversarial examples on the feature maps, we define $P_i$ as the feature maps of the original image in the $i$-th layer, $\hat{P_i}$ as the feature maps of adversarial examples, and $D(P_i||\hat{P_i})$ denotes a distance function of $P_i$ and $\hat{P_i}$. The distance between feature maps at each layer can be defined as 

\begin{eqnarray}\label{equ3}
D(P_i||\hat{P_i})=\frac{\Vert P_i-\hat{P_i} \Vert}{\Vert P_i \Vert},
\end{eqnarray}
where $||\cdot||$ represents the Euclidean norm. We also apply KL divergence \cite{joyce2011kullback} to measure the difference between the distribution of $P_i$ and $\hat{P_i}$ as
\begin{eqnarray}\label{equ4}
D_{KL}(P_i||\hat{P_i}) = \sum \mu_j(P_{ij})\log{\frac{\mu_j(P_{ij})}{\mu_j(\hat{P_{ij}})}},
\end{eqnarray}
where $P_{ij}$ represents the feature map obtained by $j$-th convolution kernel in the $i$-th layer, and $\mu_j(P_{ij}) =  \Vert P_{ij}\Vert / \sum_j^J { \Vert P_{ij} \Vert}$ is the empirical distribution through discrete samples~\cite{Boyer18221}. Then, we generate some adversarial examples on InceptionV1~\cite{GoogleNet} with several attack algorithms, including the Fast Gradient Sign Method (FGSM)~\cite{goodfellow2014explaining}, the Basic Iterative Method (BIM)~\cite{kurakin2016adversarial}, and random noise for comparison.
We randomly choose 100 classes from the ImageNet dataset~\cite{5206848}, and each class includes 200 images. The result of the experiment is shown in Figure~\ref{fig:distance}.

As shown in Figure~\ref{fig:distance}, the feature maps of adversarial examples in each layer are different from those of the original images. Moreover, the difference between the feature maps of adversarial examples and the original image becomes larger as the layer deepens, regardless of the distance metrics. This phenomenon aligns with our definition of the AFS, which was also found by Liao et al~\cite{liao2018defense}. The prediction results of the images with random noise are the same as those of the original images, and the difference of feature maps between them changes are relatively negligible even with the layers deepening. Therefore, the AFS is caused by adversarial perturbation rather than random noise. The AFS shows the changing pattern of the difference between adversarial examples and the original images through layers of different depth, substantiating its potential to be used to detect adversarial examples.

\section{Adversarial Feature Genome}

Correct classification of adversarial examples requires access to features about the original class. We believe that an input image, regardless of it being an original image or an adversarial example, demonstrates semantic features about the classes in the feedforward process of the model. Such features are hierarchical, from low-level to high-level semantic features. In this section, we propose a multi-layer data structure called the Adversarial Feature Genome (AFG), which contains the main semantic features of the input image in the feedforward process. These semantic features, like gene expressions, synthesize the final classification result from the CNN.

\subsection{Group Features}
The main components of AFG data are the semantic features about the classes in each layer of the input image under the feedforward process.
These semantic features are derived from the extraction of features by convolutional kernels. However, for each input image, it is difficult to analyze all the convolutional kernels at the same time. Olah et al.~\cite{olah2018building} propose a group visualization method, visualizing the semantic features of the input image into several important group features on each layer.
These group features contain features related to the classes, and they are also key to the formulation of the AFGs.

\begin{figure}[h]
\includegraphics[scale=1]{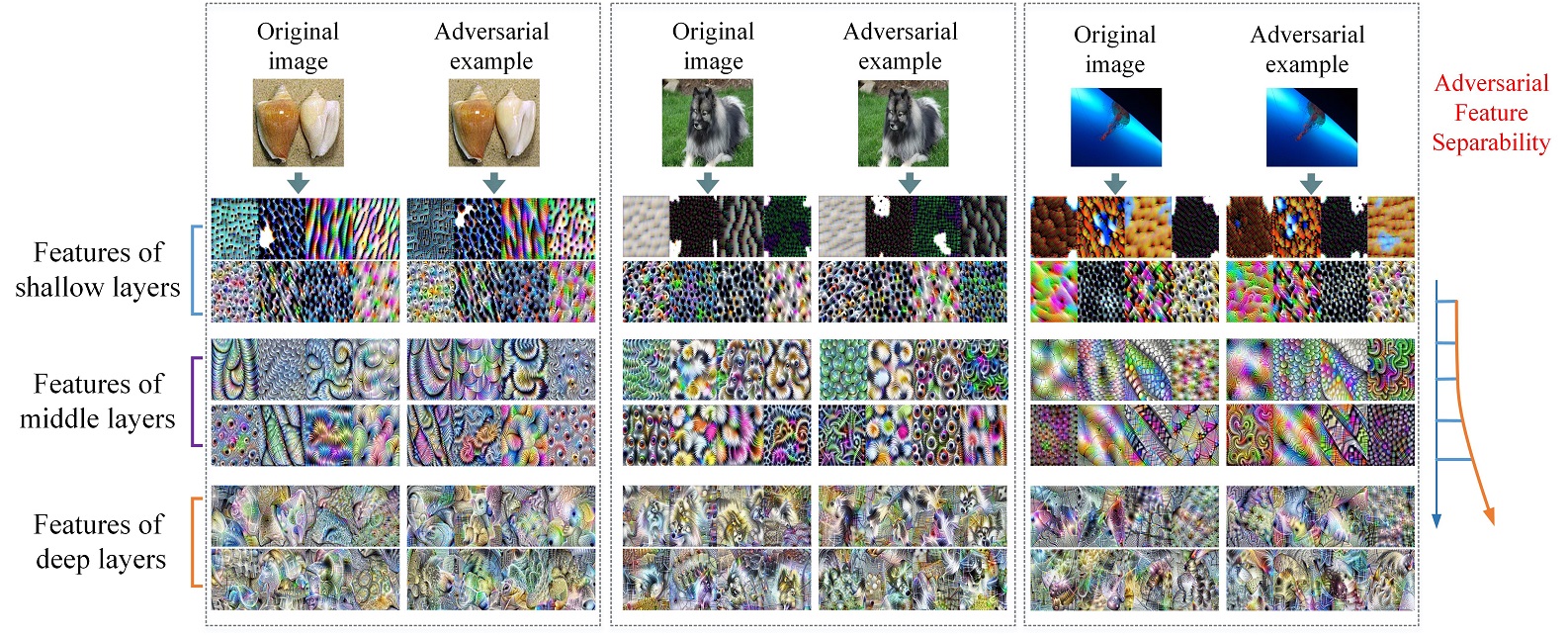}
\caption{The three sets of images and corresponding adversarial examples are conch and teddy, dog and cat, jellyfish, and paper towel. We get the group features of each input image on each layer of the model. In the shallow layers, group features of adversarial examples and original images are similar, but they differ in the deeper layers. This difference is the AFS.}
\label{fig:genome}
\end{figure}

In detail, for input image $X$, $P_i$ is the feature maps in the $i$-th layer. Considering the effect of the activation function, $P$ denotes the feature maps after the activation function. It can be expressed as $P_{i}^{W , H, C}$, where $W$, $H$, and $C$ denote the width and height of the feature maps and the number of channels. The number of channels in the feature map $P$ is the same as the number of convolution kernels. The feature map of each channel is also the response of the corresponding convolutional kernel. To obtain the main responses in these feature maps, we extract the major components of group features through matrix factorization, and in turn, we restore the semantic representations with the extracted features with gradient ascent algorithms. Specifically, we first convert $P_i$ to $P_{i}^{(W\times H), C}$, i.e., each feature map of the layer is converted into column vectors and merged into a matrix. For activation functions such as ReLU, the matrix $P_{i}^{(W\times H), C}$ is a non-negative matrix. We apply the non-negative matrix factorization to reduce dimensions and find non-negative matrix factors $U$ and $V$ such that
\begin{eqnarray}
P_{i}^{(W\times H), C} \approx (UV)^{(W\times H), C} = \sum^r_{a=1}U^{(W\times H), a} V^{a, C},
\end{eqnarray}
where $r$ represents the number of main features, which is the number of group features in the layer. Then, we apply the activation maximization algorithm \cite{Erhan2009Visualizing} on each group to obtain the group features $X_{ia}^*$ as follows:
\begin{eqnarray}
 X_{ia}^* = \sum^r_{a=1} \arg \max h_{ia}(U_{(W\times H), a} V_{a, C}, \Theta^*),
\end{eqnarray}
where $h_{ia}(X_,\Theta^*)$ is the value of the $a$-th group in the $i$-th layer, and $\Theta^*$ is the weights of CNN. 
The initial input image is random noise. With several iterations following the gradient ascent algorithm, the generated image can be made to maximize the vector of decomposed feature maps of the corresponding layers of the original image, which is a group feature.
The obtained group feature $X_{ia}^*$ also can make the maximum activation of the $a$-th group on the $i$-th layer, and its size can be customized. For instance, we set $r = 4$ and visualize 3 pairs of adversarial examples and the original image shown in Figure~\ref{fig:genome}. The input image shows semantic features about the classes on each layer. The whole process is hierarchical, and the semantics is gradually abstracted. The group features of adversarial examples and the original images are similar in the shallow layer. Although humans may not understand their semantic features in deeper layers, they are more distinct compared to shallow layers, both through visual comparison and distance metrics. We generate some group features of other images in Appendix A for references.
Through our observation, group features are similar at shallow layers while the differences grow larger when processed in deeper layers, in alignment with the AFS. These multi-layer group features can be used to construct the AFGs. 

\subsection{The AFG Data Structure}
An AFG is a composition of all group features for an image. The generation of these group features not only depends on the input image but also involves the trained CNN. As a result, the AFG potentially preserves the interrelationship between the model and the original image. Therefore, the AFG has the AFS property while retaining the main semantic features about the class on each layer. All these characteristics are beneficial for the AFG to be used to detect and correctly classify adversarial examples.
Considering that the group features can show the property of the AFS only through multiple layers, the AFG should also be a multi-layer data structure. Accordingly, multiple group features are collected at each layer, and we stack the group features of all layers. These group features of the layer are also interrelated. So, we use a gene-like approach to stitch them together~\cite{weijtens1998retroviral}.

Specifically, with an input image $X$, we get $r$ group features on each layer. For instance, we set $r = 4$, and the size of group feature $X^*_{ij}$ is $112 \times 112$.
On each layer, the input image gets four $112 \times 112$ group features. We stitch four group features together on the same layer to get a larger feature $224\times 224$. Then the large features of all layers that become a tensor $224\times 224\times N$ are stacked together. $N$ represents the number of layers. This tensor is the AFG of the input image.

\section{Data-driven Adversarial Examples Recognition Framework}

AFG is a multi-layered and complex data structure. The recognition model over the AFG also needs to have good feature extraction ability. Therefore, we still use CNN to process the AFG data. On the other hand, the complex structure and the generation strategy of the AFGs allow it to avoid secondary attacks, even for training CNNs. The subsequent recognition model detects adversarial examples and predicts the original classes with the AFGs, which corresponds to two classification tasks, respectively.
It is a two-path AFG recognition model. Each classification produces a prediction for one corresponding label of the AFG.
Consistently, each AFG has two labels associated with it. One label marks the correct original class, and the other is the binary indicator of being from an adversarial example. So accurate predictions of the labels allow reliably detect and reclassify adversarial examples to the original classes simultaneously. We refer to this process as a data-driven adversarial example recognition framework.

\begin{figure}[h]
\begin{center}
\includegraphics[scale=0.98]{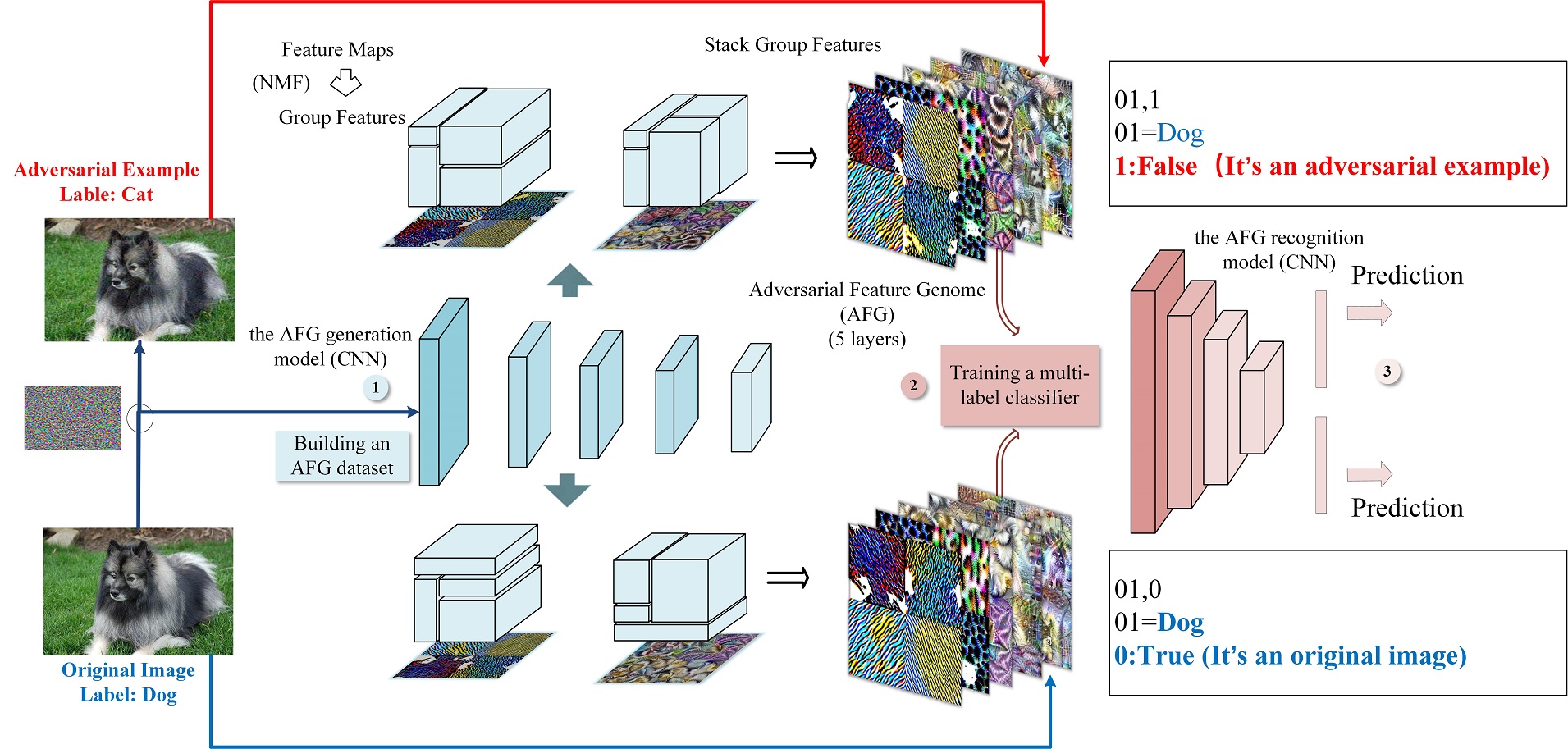}
\end{center}
\caption{A data-driven adversarial example recognition framework. With the trained AFG generation model, the detection and classification procedures in this framework can be divided into three steps. (1) We use the AFG generation model CNN to convert the input image into group features via the group visualization method. (2) We stack all stitched group features to create an AFG for each input image. The number of AFG layers is the same as the number of convolution layers. We construct a dataset from AFGs of all images. (3) The adversarial example problem is transformed into a classification task. The second two-path AFG recognition model is trained on the AFG dataset. It can detect whether the input image is an adversarial example and predict the original class of adversarial example.}

\label{fig.frame}
\end{figure}

The framework constitutes two models, as shown in Figure~\ref{fig.frame}. The first model is an AFG generation model, which is trained on the original images for a sore classification task. Adversarial examples are generated on the converged AFG generation model with the different attack algorithms. Then, we retrieve the AFGs of all original images and adversarial examples via the AFG generation model. All AFGs collectively form a new dataset. Having labeled all AFGs by the approach described above, we can train a two-path AFG recognition model on that dataset, noted as the AFG recognition model. 
We can choose from a variety of backbones for the two-path AFG recognition model. The extracted features are passed through two classifiers to determine their predicted labels for original class and adversarial example identification, achieving simultaneous detection and classification of adversarial examples.

\section{Experiments}
\subsection{Preliminary Experiments}
The data-driven adversarial example recognition framework requires two models. The first model is used for AFG generation, which is at risk from adversarial examples. The second model is used for the detection and correct classification of adversarial examples. For the AFG generation model, we use multiple configurations as our benchmark to validate the effectiveness of the AFG, including three CNNs, three datasets, and four attack algorithms for a sum of 36 attack scenarios. These CNNs are VGG16~\cite{SimonyanZ14a}, InceptionV1~\cite{szegedyLJSRAEVR15}, and ResNet50~\cite{he2016deep}, which are all commonly used models in computer vision tasks. The three datasets are Flower~\cite{NilsbackZ08}, Caltech101~\cite{LiFP04}, and Caltech256~\cite{2007Caltech}. The Flower dataset has as many as 102 classes of flowers, and each class has approximately 20-30 images. Caltech101 has a total of 101 classes. Caltech256 is an upgraded version of Caltech101, which encompasses 256 classes. The performance of these CNNs evaluated on different datasets is shown in Table~\ref{tab.benchmark}. These models perform well on these datasets. For the AFG recognition model, we use a CNN as well, whose backbone is VGG16.

\begin{table}[h]
\caption{Accuracy performance of CNNs on different datasets and the effectiveness of different attack algorithms. Two values are present for the effectiveness of each attack algorithm, one indicating the accuracy of the model after being attacked and the other value in parentheses indicating the fooling rate. (\%)}
\label{tab.benchmark}
\setlength{\tabcolsep}{1.8mm}{
\begin{tabular}{llccccc}
\hline
Dataset & Model & Accuracy & FGSM & BIM & DeepFool & C\&W \\ \hline
 & VGG16 & 90.44 & 79.80\quad(11.76) & 79.8\quad(11.76) & 0.0\quad(100.0) & 74.55\quad(17.65) \\
 & InceptionV1 & 91.73 & 37.02\quad(59.64) & 0.0.0\quad(100.0) & 0.0\quad(100.0) & 10.14\quad(88.97) \\
\multirow{-3}{*}{Flower} & ResNet50 & 88.79 & 75.57\quad(14.89) & 75.32\quad(15.17) & 0.0\quad(100.0) & 64.29\quad(27.67) \\ \hline
 & VGG16 & 66.08 & 41.67\quad(36.89) & 41.62\quad(36.96) & 0.0\quad(100.0) & 36.21\quad(45.19) \\
 & InceptionV1 & 92.96 & 31.69\quad(65.88) & 0.17\quad(99.82) & 0.07\quad(99.93) & 5.74\quad(93.82) \\
\multirow{-3}{*}{Caltech101} & ResNet50 & 78.70 & 53.59\quad(31.47) & 52.14\quad(33.33) & 0.0\quad(100.0) & 37.77\quad(51.76) \\ \hline
 & VGG16 & 71.29 & 46.23\quad(35.14) & 45.98\quad(35.5) & 0.0\quad(100.0) & 34.11\quad(52.16) \\
 & InceptionV1 & 78.71 & 7.90\quad(89.96) & 0.0.0\quad(100.0) & 0.0\quad(100.0) & 0.86\quad(98.91) \\
\multirow{-3}{*}{Caltech256} & ResNet50 & 78.32 & 51.0\quad(34.56) & 48.23\quad(38.11) & 0.0\quad(100.0) & 30.28\quad(61.14) \\ \hline
\end{tabular}}
\end{table}

Four attack algorithms are applied to fool the AFG generation model, FGSM~\cite{goodfellow2014explaining}, BIM~\cite{kurakin2016adversarial}, DeepFool~\cite{moosavi2016deepfool}, C\&W~\cite{carlini2017towards}. The results of the attacks are also shown in Table~~\ref{tab.benchmark}. There are two metrics for the effectiveness of attack algorithms. The first indicates the accuracy of the model after being attacked, and the second (in parentheses) indicates the fooling rate of the attack algorithm. A high fooling rate suggests a successful attack, and the accuracy of the model is consequently lower. All these attack algorithms undermine the accuracy of the base classification model. Notably, DeepFool invokes the strongest attacks among these classification tasks, fooling the base classification models almost entirely. The FGSM attack algorithm is the weakest among all the attack algorithms, as is presented in the results. 

In the next step, we retrieve all AFGs in these 36 attack scenarios, along with AFGs of the original images. We run 100 iterations of gradient ascend for every group feature generation. The AFGs generated are collected as the training set for the AFG recognition model to detect and reclassify adversarial examples.

\subsection{The AFS Property of AFGs}

The AFS represents the pattern of difference between the features interpreted by the CNN from the original image and adversarial examples. We can take advantage of it to detect adversarial examples. Yet, that the AFGs carry the AFS property remains a hypothesis until we verify it. Only if this hypothesis is accepted shall AFGs be used to distinguish adversarial examples.

\begin{figure}[ht]
\begin{center}
\includegraphics[scale=0.71]{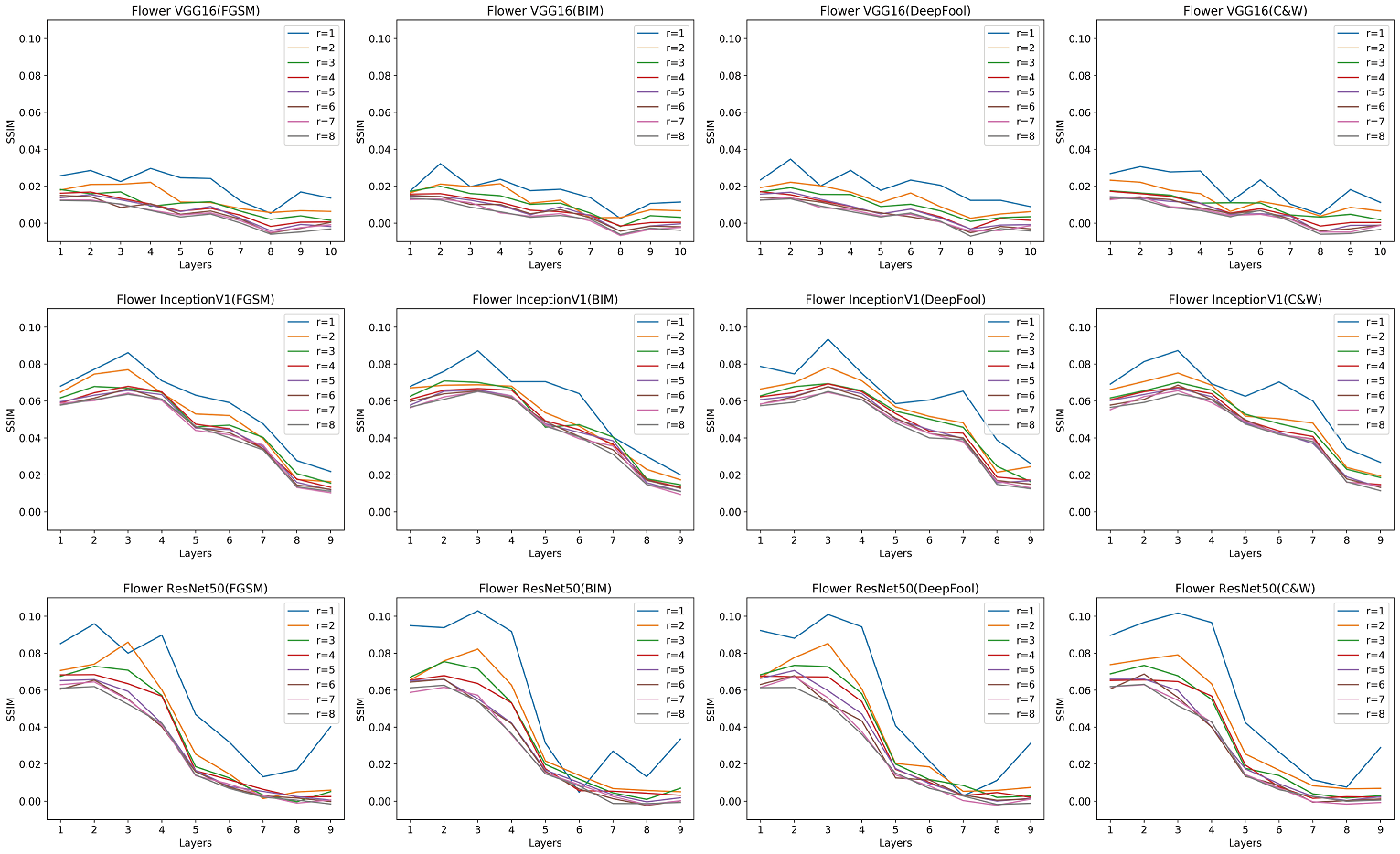}
\end{center}
\caption{Layer-by-layer differences in the AFGs of the original images and adversarial examples for the 12 attack scenarios under the Flower dataset. $r$ indicates the number of group features per layer that lead to different structures of the AFGs. And they all show a tendency for the differences to become larger as the number of layers deepens.}
\label{fig.ars}
\end{figure}

According to the AFG data structure definition, the AFG generation model generates $r$ sets of group features at each convolutional layer for each image. To verify that the AFG has the AFS property, we set multiple sets of $r$, from 1 to 8. Unlike feature maps, the order of these group features are not aligned. Thus, it is inappropriate to use the same distance evaluation metrics as we do with the feature map. On the other hand, many textures are similar, but their Euclidean distances are the difference between the corresponding positions, which can be large. Therefore, we use the structural similarity index (SSIM)~\cite{1284395} to assess the difference between the AFGs of an original image and the corresponding adversarial example at corresponding layers. The value of SSIM ranges from $[-1,1]$, with 1 indicating that they are most similar.
The SSIM at a layer is retained by calculating the average pairwise similarity from the group features over the original image to the group features over the adversarial example at the same layer.

The experimental results of 12 attack scenarios under the Flower dataset are shown in Figure~\ref{fig.ars}. In every attack scenario, the similarity between the AFG of the original image and the corresponding layer of the adversarial example's AFG exhibits a decreasing trend as the features go through deeper layers. This indicates that the difference between the AFG is also getting larger. This phenomenon is consistent with the AFS property. We apply an array of $r$  parameter settings to examine the effect of the number of group features have on the similarity metric. We find that when $r = 1$, the trend of the similarity fluctuates drastically as the layer deepens, especially for ResNet50. However, this fluctuation is not observed when the number of the group features is larger, which is consistent with our hypothesis of the AFS. Therefore, in the subsequent experiments, we dictate a standard AFG structure. The standard AFG has 4 group features in each layer, and the size of each group feature is $112\times 112$. Experimental results for the other two datasets are in Appendix B. They also have the same findings. These experimental results demonstrate that AFGs carry the property of the AFS.

\subsection{Adversarial Example Detection}

To verify whether the AFS property of AFGs can be exploited for adversarial example detection, we remove the classification path in our two-path AFG recognition model to focus sorely on detecting adversarial examples.
This leaves us with an adversarial example detection model, which can be compared with other defense algorithms.
We choose the LID as the baseline for comparison, and we find that the AFG recognition model has demonstrated a significant edge in terms of the accuracy of the detection. The experimental results are shown in Table~\ref{tab.defense}.

\begin{table}[h]
\caption{Accuracy of adversarial example detection in different attack scenarios. (\%)}
\label{tab.defense}
\setlength{\tabcolsep}{2.7mm}{
\begin{tabular}{ll|cccccccc}
\hline
 &  & \multicolumn{2}{c}{FGSM} & \multicolumn{2}{c}{BIM} & \multicolumn{2}{c}{DeepFool} & \multicolumn{2}{c}{C\&W} \\
\multirow{-2}{*}{Dataset} & \multirow{-2}{*}{Model} & LID & AFG & LID & AFG & LID & AFG & LID & AFG \\ \hline
 & VGG16 & 76.08 & 98.02 & 71.79 & 99.01 & 70.17 & 92.58 & 69.58 & 92.13 \\
 & InceptionV1 & 82.58 &  92.61 &  77.75 &  93.18 &  66.67 & 91.01 & 66.67 & 90.79 \\
\multirow{-3}{*}{Flower} & ResNet50 & 80.92 & 90.01 & 76.67 & 89.54 & 68.13 & 90.07 & 66.67 & 88.06 \\ \hline
 & VGG16 & 81.26 & 74.70 & 76.99 & 74.41 & 76.29 & 74.44 & 74.63 & 71.88 \\
 & InceptionV1 & 81.67 & 82.26 & 81.78 & 81.41 & 72.54 & 85.88 & 66.67 & 76.70 \\
\multirow{-3}{*}{Caltech101} & ResNet50 & 85.68 & 79.01 & 80.47 & 78.51 & 72.71 & 79.43 & 70.69 & 77.12 \\ \hline
 & VGG16 & 80.87 & 71.39 & 75.25 & 70.35 & 70.58 & 71.25 & 70.92 & 71.03 \\
 & InceptionV1 & 80.08 & 82.31 & 78.25 & 91.38 & 66.67 & 76.49 & 66.67 & 77.38 \\
\multirow{-3}{*}{Caltech256} & ResNet50 & 79.67 & 90.89 & 73.42 & 91.37 & 66.71 & 81.42 & 66.67 & 84.89 \\ \hline
\end{tabular}}
\end{table}

The average accuracy of the adversarial example detection model with the AFGs is 88.17\%. Among all three datasets, the detection model performs best in the Flower dataset. Against the four attack algorithms, our detection model achieves average accuracies of 93.55\%, 93.91\%, 91.22\%, and 90.32\%, respectively. And it performs best in detecting adversarial examples from FGSM and BIM attack algorithms. The AFG-based detection model also outperforms the LID in most attack scenarios, especially against DeepFool and C\&W algorithms. The AFG has an average edge in the accuracy of 9.38\% in all attack scenarios against the LID. These experiments demonstrate that the AFG with the AFS property can be directly used for adversarial example detection.

\subsection{The Different Layers of AFG}\label{sec.layer}

We further analyzed the reasons why the AFG can be used for adversarial example detection. We believe that the AFG with a hierarchical structure contains cues about classes on each layer. When this cue about the class is inconsistent as the layer changes, it may be detected as an adversarial example. In this regard, we test this hypothesis by using AFGs with different layers.
We gather a large dataset with AFGs of different attack scenarios under the same dataset. Then, we excerpt three sets of three-layer AFGs, from the shallow, middle, and deep layers of full-size AFGs, defined as AFG-S, AFG-M, and AFG-D.
The new excerpted AFGs have the same structure as the full-size ones, though their layer size is not constrained by the number of layers in the AFG generation model. The detection model trained with AFGs over one dataset can also be transferred to detect adversarial examples from other datasets. The experimental results are shown in Table~\ref{tab.afg-s}, Table~\ref{tab.afg-m}, and Table~\ref{tab.afg-d}.

\begin{table}[h]
\caption{Accuracy and Transferability of AFG-S based adversarial example detection model.  (\%)}
\label{tab.afg-s}
\setlength{\tabcolsep}{11mm}{
\begin{tabular}{cccc}
\hline
 &  Flower& Caltech101& Caltech256 \\ \hline
 Flower& 92.36& 68.14& 71.17 \\
 Caltech101& 81.75& 77.56& 74.45 \\
 Caltech256& 80.46& 73.23& 82.70 \\ \hline
\end{tabular}}
\end{table}

\begin{table}[h]
\caption{Accuracy and Transferability of AFG-M based adversarial example detection model.  (\%)}
\label{tab.afg-m}
\setlength{\tabcolsep}{11mm}{
\begin{tabular}{cccc}
\hline
 &  Flower& Caltech101& Caltech256 \\ \hline
 Flower& 89.17& 67.58& 64.64 \\
 Caltech101& 73.52& 78.54& 66.15 \\
 Caltech256& 63.86& 61.66& 83.80 \\ \hline
\end{tabular}}
\end{table}

\begin{table}[h]
\caption{Accuracy and Transferability of AFG-D based adversarial example detection model.  (\%)}
\label{tab.afg-d}
\setlength{\tabcolsep}{11mm}{
\begin{tabular}{cccc}
\hline
 &  Flower& Caltech101& Caltech256 \\ \hline
 Flower& 80.05& 62.64& 62.42 \\
 Caltech101& 65.84& 68.87& 65.66 \\
 Caltech256& 60.81& 59.52& 75.61 \\ \hline
\end{tabular}}
\end{table}

The experiments show that regardless of which set of AFGs is used to train detection models, they can accurately detect adversarial examples while showing good transferability. The adversarial example detection models trained on one dataset can also detect adversarial examples on other datasets, even if they have different semantic classes. This demonstrates the generality of the detection models trained with the AFGs. We also find that the detection models trained with the AFG-S set have the best performance in accuracy and transferability. Per the pattern of AFG differences in Figure~\ref{fig.ars}, AFG-D has the largest difference between original images and adversarial examples. However, this is contradictory to the performance of the detection model for adversarial examples. We believe that Figure~\ref{fig.ars} demonstrates the trend of the variation of the features about the classes. The images implicitly have consistent cues about the features of the classes on each layer of the feedforward operation. And this consistent cues about adversarial examples evolves intensely at the shallow layer, i.e., the change rate in the differences are higher than that of deeper layers. The changing pattern of differences between the original image and adversarial examples along ascending levels of layers through CNN is perhaps the essence of the AFS property.

\subsection{The Features about Classes}

The AFS property of AFGs is used to detect adversarial examples, while the reclassification of adversarial examples requires that AFGs embody features about the classes.
For this, we update the labels of AFGs of all original images and set their labels only to the labels of the original classes.
And we focus on the one path of the AFG recognition model, which is the multi-class classification model.
The experimental results are shown in Table~\ref{tab.mulclass}.

\begin{table}[h]
\caption{Accuracy of InceptionV1 under different datasets of original images and corresponding AFG.  (\%)}
\label{tab.mulclass}
\setlength{\tabcolsep}{12mm}{
\begin{tabular}{cccc}
\hline
 & Flower & Caltech101 & Caltech256 \\ \hline
Images & 91.73 & 92.96 & 78.71 \\
AFG & 90.26 & 90.93 & 75.16 \\ \hline
\end{tabular}}
\end{table}

The AFGs are derived from the original images and the classification model in the AFG generation model. Compared to the base classification model trained under the original image, the accuracy of the AFG recognition model trained with AFGs is reduced by 0.47\%, 2.03\%, and 2.55\% on our three datasets. Though this reduction is slight, and the AFG recognition model still demonstrates satisfactory classification performance. This also shows that the hierarchical structure of the AFG contains features about the classes. It is the key to reclassify adversarial examples. 

\subsection{Adversarial Example Recognition}

Carrying both AFS property and features about classes, the AFG recognition model can simultaneously detect and correctly classify adversarial examples. We use the standard data-driven adversarial example recognition framework. The AFG recognition model, which is a two-path AFG recognition model, produces two classification results. One is about the semantic class of the input AFG, with no regard to whether being from the adversarial example. Meanwhile, the other classifier is used to detect the adversarial example. The experimental results are shown in Table~\ref{tab.rec}.

On the three datasets, their average accuracy is 84.60\%, 75.25\%, and 59.53\%. The accuracy of the two-path AFG recognition model refers to the intersection over the union of the two classifiers' results with the ground truth. On the Flower dataset, the VGG16 performs the worst among models regardless of the attack algorithms, while the InceptionV1 is the best. The accuracy of VGG16 is also low on the other two datasets. This may be because the features learned by inception are more semantically meaningful compared to VGG16~\cite{olah2017feature}.
The features learned by VGG16 about classes are more easily fooled, which also makes it easier to transfer adversarial examples generated under VGG16 to other models~\cite{8099500}.
The difference in models' accuracies against the four attack algorithms is not significant when compared with the apparent difference in fooling rates shown in Table~\ref{tab.benchmark}.
DeepFool scenarios, which have the largest fooling rate against the base semantic classification model, also yield similar defense performance to attack FGSM. However, on the Caltech101 dataset, the defense performance decreases for all attack scenarios compared to the performance in the Flower dataset. This finding also applies to the Caltech256 dataset with more classes. On the Caltech256 dataset, the model has accomplished the worst defensive performance amongst all datasets. This indicates that the complexity of datasets affects defense performance. However, all these experimental results demonstrate that recognition models trained with AFGs can reliably fulfill the goal to detect and classify adversarial examples simultaneously.

\begin{table}[h]
\caption{Performance of adversarial example recognition framework under different attack scenarios. (\%)}
\label{tab.rec}
\setlength{\tabcolsep}{6.5mm}{
\begin{tabular}{llcccc}
\hline
Dataset & Model & FGSM & BIM & DeepFool & C\&W \\ \hline
 & VGG16 & 79.70 & 78.93 & 79.50 & 79.87 \\
 & InceptionV1 & 94.07 & 92.95 & 92.30 & 83.97 \\
\multirow{-3}{*}{Flower} & resnet & 86.75 & 83.65 & 82.79 & 80.73 \\ \hline
& VGG16 & 66.86 & 68.25 & 77.60 & 64.63 \\
 & InceptionV1 & 87.50 & 75.00 & 76.88 & 72.94 \\
\multirow{-3}{*}{Caltech101}  & resnet & 87.02 & 80.38 & 85.50 & 80.46 \\ \hline
 & VGG16 & 62.08 & 73.32 & 50.03 & 51.20 \\
 & InceptionV1 & 70.83 & 65.00 & 59.05 & 56.00 \\
\multirow{-3}{*}{Caltech256} & resnet & 70.42 & 86.67 & 56.79 & 62.94 \\ \hline
\end{tabular}}
\end{table}

An input image is associated with three predictions through the framework, including the prediction from the base classification model and the two additional predictions from the AFG recognition model. When the three output results agree, we can confirm the security and classification results of the input image. Otherwise, the framework is either inaccurate for this image or under attack. This inspires us to filter out the possible wrong predicted images for new data without labels, guaranteeing the results.

For AFGs with a multi-layered structure, we also extract different layers of AFGs to explore the effect of information captured at different layers. In the same configuration as the previous Experiment~\ref{sec.layer}, we use AFG-S, AFG-M, and AFG-D on the Flowers dataset to verify the defensive performance against the attack algorithm. The experimental results are shown in Table~\ref{tab.layer}.

\begin{table}[h]
\caption{Performance of adversarial example recognition of AFG with different structures under Flower dataset. (\%)}
\label{tab.layer}
\setlength{\tabcolsep}{9.5mm}{
\begin{tabular}{lcccc}
\hline
 & AFG-S & AFG-M & AFG-D & AFG \\\hline
FGSM & 43.89 & 73.87 & 93.45 & 94.07 \\
BIM & 42.96 & 73.98 & 92.26 & 92.95 \\
DeepFool & 43.39 & 69.51 & 82.45 & 83.97 \\
C\&W & 51.19 & 70.41 & 75.33 & 92.30 \\ \hline
\end{tabular}}
\end{table}

The recognition model trained with the AFG-S has the worst defensive performance for all four attack algorithms, while the recognition model trained with AFG-D has the best performance. 
This experimental conclusion is opposite to the results in Experiment~\ref{sec.layer}.
However, as can be seen from Table~\ref{tab.layer}, the accuracy of the adversarial example detection models trained with AFG-S and AFG-D does not differ much compared to the defense performance under different layers of the AFG demonstrated in Table~\ref{tab.afg-s} and Table~\ref{tab.afg-d}. This may attribute to the fact that the deeper-layer AFGs contain more semantic features about the classes. At the same time, the AFS property used for adversarial example detection exists from shallow to deep layers.
In summary, shallow AFGs can be better used to detect adversarial examples, while deep AFGs are better for reclassifying adversarial examples.

\subsection{The Transferability of Framework}

Unlike the detection of adversarial examples with a binary classification task, the results of reclassifying adversarial examples are the same as the number of classes in the corresponding dataset. We cannot do the transferable analysis between datasets.
So we explore multiple attack scenarios with the Flower dataset. The adversarial example detection model against one attack algorithm is directly used to defend against other attack algorithms, and the experimental results are shown in Table~\ref{tab.transre}.

\begin{table}[h]
\caption{Transferability of adversarial example recognition for different attack algorithms with the Flower dataset. (\%)}
\label{tab.transre}
\setlength{\tabcolsep}{8mm}{
\begin{tabular}{l|c|ccc}
\hline
& Source & \multicolumn{3}{c}{Target} \\ \cline{2-5} 
\multirow{-1}{*}{Model}  & FGSM & BIM & DeepFool & C\&W \\ \hline
VGG16 & 94.07 & 93.94 & 90.22 & 64.44 \\
InceptionV1 & 79.70 & 74.42 & 64.58 & 63.01 \\
ResNet50 & 86.75 & 86.69 & 76.72 & 68.28 \\ \hline
 & BIM & FGSM & DeepFool & C\&W \\ \hline
VGG16 & 92.95 & 93.15 & 89.14 & 63.73 \\
InceptionV1 & 78.93 & 74.64 & 62.94 & 63.19 \\
ResNet50 & 83.65 & 83.87 & 74.20 & 66.09 \\ \hline
 & DeepFool & FGSM & BIM & C\&W \\ \hline
VGG16 &83.97 & 82.75 & 82.75 & 80.81 \\
InceptionV1 &80.73 & 70.77 & 59.51 & 81.00 \\
ResNet50 &79.87 & 81.93 & 81.96 & 78.27 \\ \hline
 & C\&W & FGSM & BIM & DeepFool \\ \hline
VGG16 &92.30 & 91.24 & 95.28 & 72.04 \\
InceptionV1 &82.79 & 72.96 & 61.67 & 82.58 \\
ResNet50 &79.50 & 85.76 & 85.75 & 72.25 \\ \hline
\end{tabular}}
\end{table}

All these recognition models also show good transferability. In some attack scenarios, the accuracy of the transferred recognition models is even higher than that of the related origin model. The attack algorithms have a similar impact on the models, even though they have different strategies. These experimental results demonstrate that the framework has good generalization over different attack algorithms.

\subsection{The Different Recognition Models}

\begin{table}[h]
\caption{Performance of two-path AFG recognition models with different architectures under the Flower dataset. (\%)}
\label{tab.classifier}
\setlength{\tabcolsep}{9mm}{
\begin{tabular}{lcccc}
\hline
 & FGSM & BIM & DeepFool & C\&W \\ \hline
VGG16 & 94.07 & 92.95 & 92.30 & 83.97 \\
InceptionV1 & 93.51 & 93.51 & 91.56 & 84.36 \\
ResNet50 & 90.88 & 91.59 & 86.33 & 79.98 \\ \hline
\end{tabular}}
\end{table}

The second model of adversarial example recognition framework is also a CNN, and we apply multiple architectures to demonstrate their performance on the attack algorithms under the Flower dataset. The experimental results are shown in Table~\ref{tab.classifier}. We find that the second model performs differently under various attack algorithms. VGG16 gives the best results in defense against the FGSM. InceptionV1 has better performance against the BIM attack algorithm. Therefore, the architecture of the second model is also an important choice in the adversarial examples recognition framework.

\section{Discussion and Conclusion}
With the widespread deployment of CNNs, the risk posed by adversarial examples is becoming increasingly serious. Therefore, it would benefit the system's robustness to defend successfully and even reclassify adversarial examples correctly. We need to find their differences from the original images to detect adversarial examples, which we call AFS. In order to reclassify adversarial examples, we need to discover features on adversarial examples with respect to the original classes. Based on these two ideas, we propose the AFG, which contains both the property of AFS and the features of the original classes.

We believe that an adversarial example recognition model trained on AFGs can defend against different attacks. We used a large number of configurations as benchmarks to validate the effectiveness of AFGs.
First, we generated multiple AFGs with different structures, i.e., different numbers of group features. Then the differences between the AFG of the original image and that of adversarial examples both become larger as the number of layers deepens, regardless of the structure of the AFG. The AFG has the AFS property. Then we give AFG a binary classification label indicating whether it is from an adversarial example and use them for training an adversarial example detection model. The experiments show that the adversarial example detection model performs well in a variety of configurations of attack scenarios. We further explore the detection accuracy and transferability of AFG with different layers. Based on the experimental results, we explain that the reason why AFGs can be used for the detection of adversarial examples is that hierarchical AFGs have clues about the consistency of the same class. When this consistency has a difference, it is detected as an adversarial example. This may be the essence of AFS. On the other hand, we verify whether the AFG has features about the classes. For this, we trained the classification models directly on the AFG of the original image. Their accuracy is not much different from the accuracy of the original image classification model. The AFG has features about classes. Thus, AFGs with AFS as well as features about classes can be used to detect and reclassify adversarial examples simultaneously. Then, based on our proposed adversarial example recognition framework, the two-path AFG recognition model performs well in several attack scenarios. They all successfully detect and classify the adversarial examples. We also explored the transferability of these recognition models and the use of different layers of AFG. They all show good results. While adversarial examples in the transferability experiments are unknown for our defense model, these results show that the two-path AFG recognition model has good defense capability in black-box settings.
Finally, we use two-path AFG recognition models with different architectures. The recognition model has different performances for different attacks. 
We also validate our defense framework on black-box attacks~\cite{tu2019autozoom}. Referring to Appendix C for detailed results, the proposed framework still recovers the correct labels of adversarial examples under the black-box attack.
These experiments validate the effectiveness of AFGs and provide a new data-driven strategy for defending against adversarial examples.

AFG also has limitations. Each AFG comes from a transformation of the input image, which is time-consuming and computationally complex. Regarding the structure of AFG, our approach of stitching and stacking group features together may not be the best strategy. In particular, regarding the AFS property, although we give an explanation, it is still an unsolved problem.
More, AFG also has many problems that need to be explored. For example, shallow AFG can better detect adversarial examples, while deep AFG can better classify them. And the quantitative analysis of the differences between AFG. More discussion and validation on these issues are needed. These are also the topics of our future research.

\section*{ACKNOWLEDGEMENTS}\label{ACKNOWLEDGEMENTS}

This work was supported by the National Natural Science Foundation of China (grant numbers 41871364, 41871276, 41871302, and 41861048). This work was carried out in part using computing resources at the High Performance Computing Center of Central South University.

\bibliography{sample}

\newpage

\appendix
\section{Group Features of other Images}\label{other-gp}

\begin{figure}[h]
\centering
\includegraphics[scale=0.98]{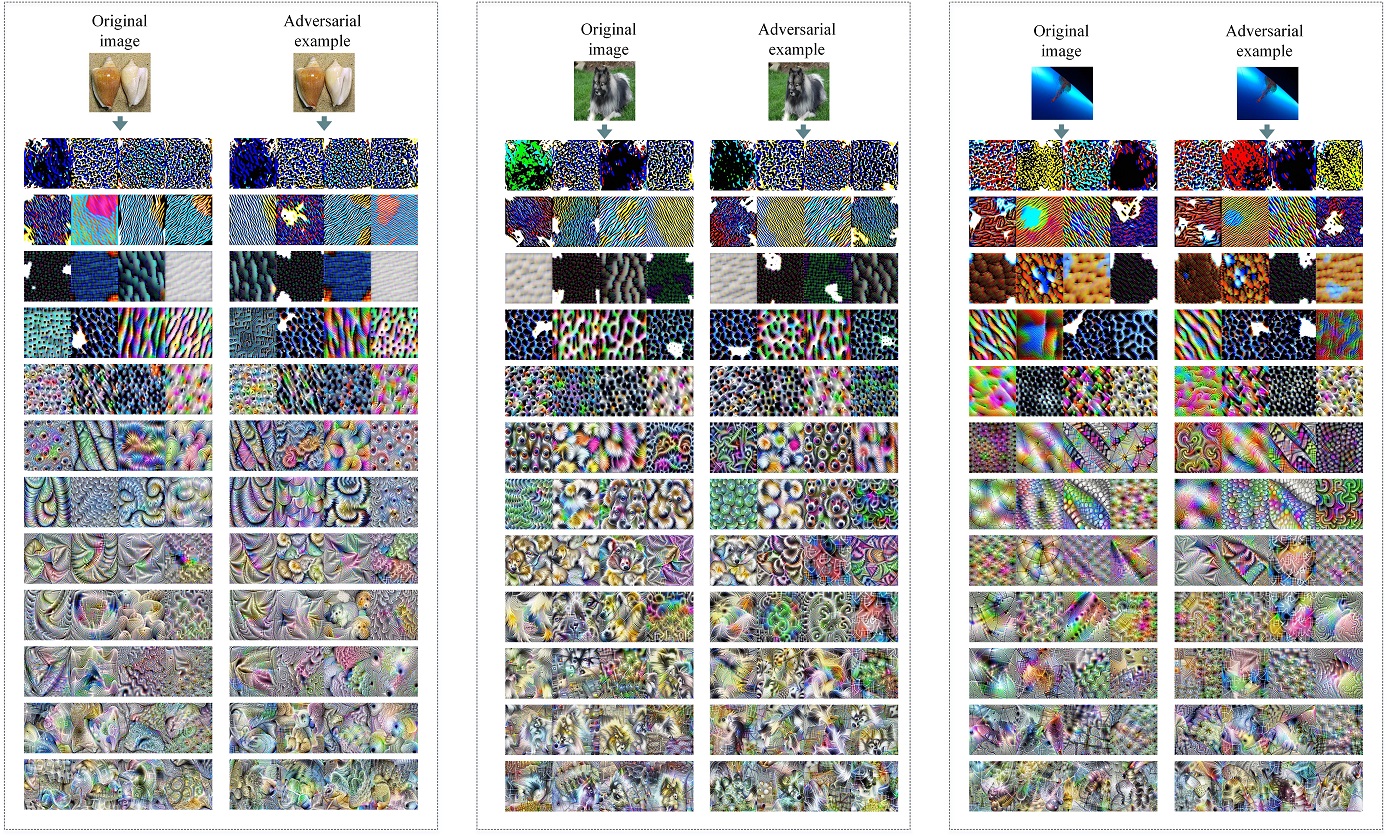}
\caption{The three sets of images and corresponding adversarial examples are conch and teddy, dog and cat, jellyfish, and paper towel.}
\end{figure}

\begin{figure}[h]
\centering
\includegraphics[scale=0.98]{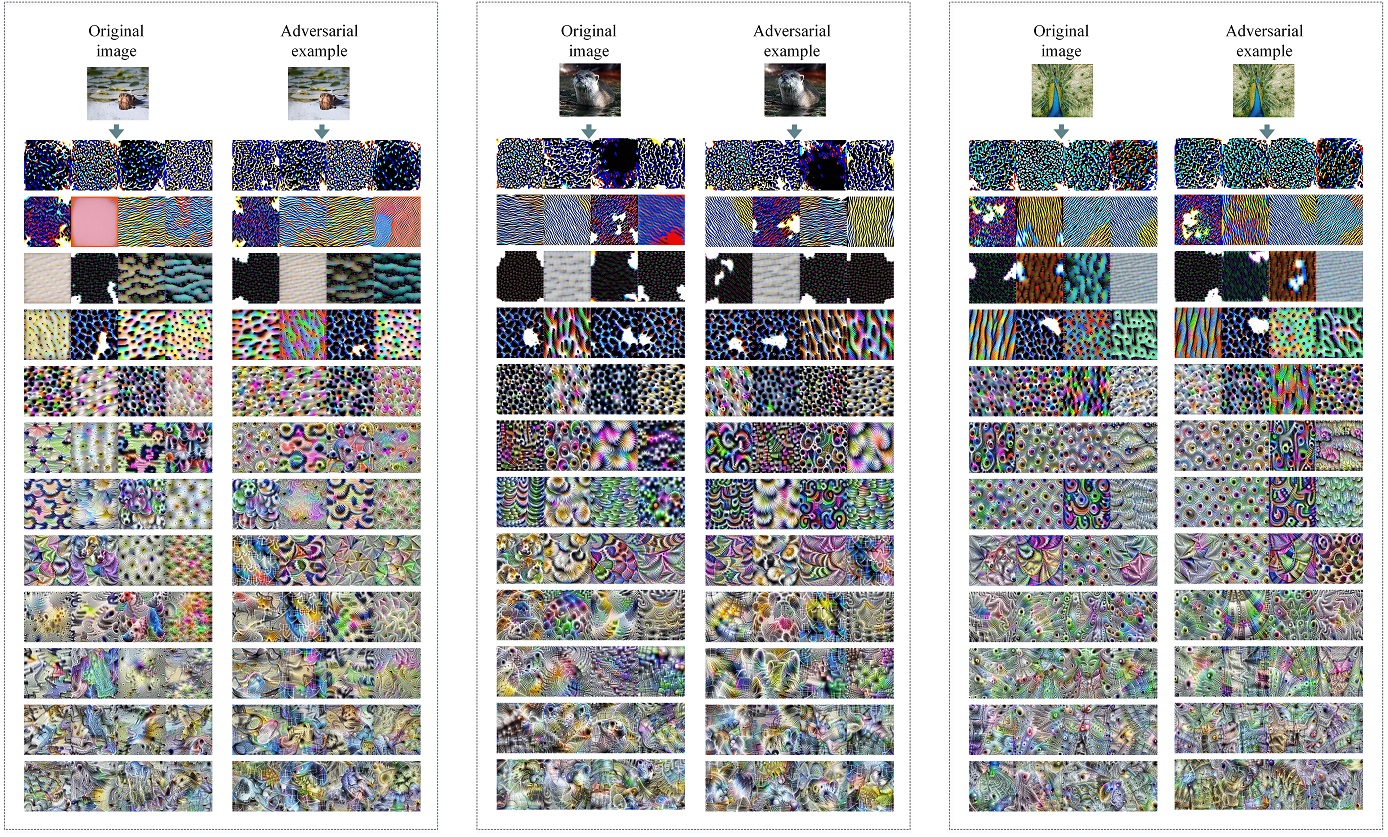}
\caption{The three sets of images and corresponding adversarial examples are beaver and cowboy hat, otter and keeshond, peacock, and cardigan.}
\end{figure}

\newpage
\section{The AFS in Other Datasets}\label{other-afs}

\begin{figure}[htb]
\begin{center}
\includegraphics[scale=0.51]{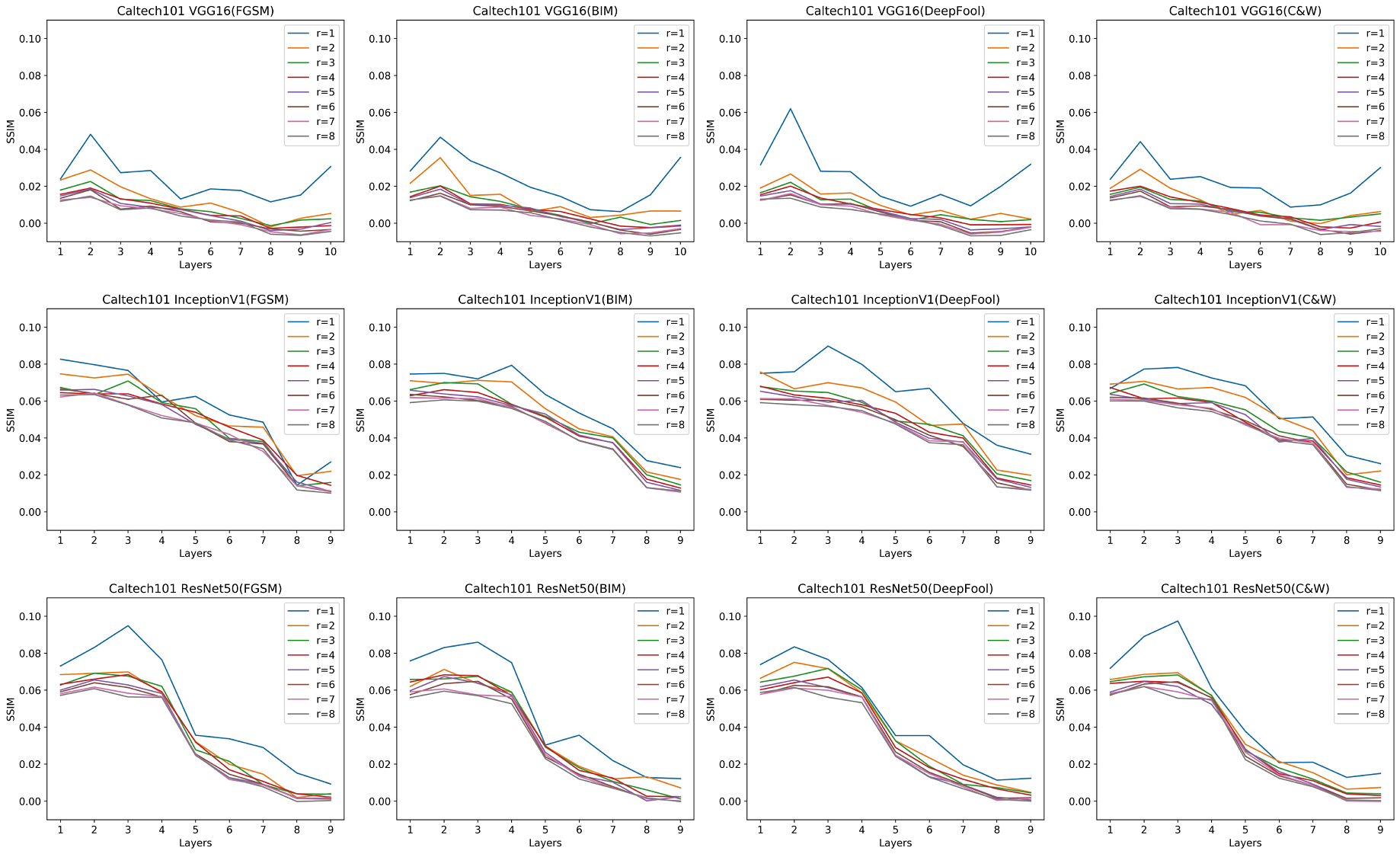}
\end{center}
\caption{Layer-by-layer differences in the AFG of the original images and adversarial examples for the 12 attack scenarios under the Caltech101 dataset.}
\end{figure}

\begin{figure}[htb]
\begin{center}
\includegraphics[scale=0.51]{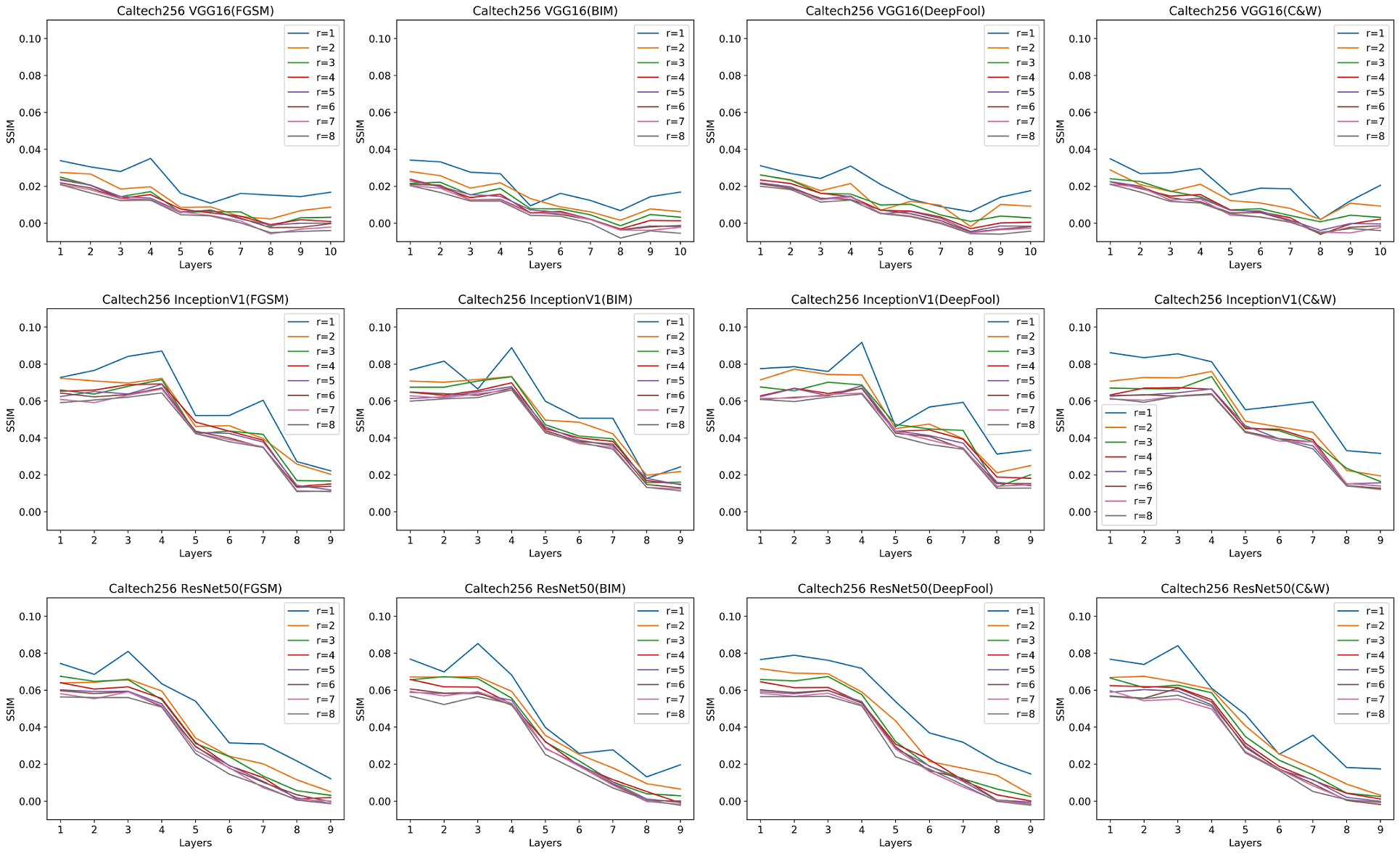}
\end{center}
\caption{Layer-by-layer differences in the AFG of the original images and adversarial examples for the 12 attack scenarios under the Caltech256 dataset.}
\end{figure}

\newpage
\newpage

\section{Performance on Black-Box Attacks}\label{other-black}
To validate the framework's effectiveness in black-box attacks, we use multiple CNNs on the Flower dataset, including VGG16, InceptionV1, and ResNet50. We choose AutoZOOM~\cite{tu2019autozoom} as the black-box attack algorithm for generating adversarial examples, which is a black-box attack using random full gradient estimation and data-driven acceleration. AutoZOOM uses adaptive random gradient estimation and dimension reduction techniques to reduce the attack query counts while maintaining attack effectiveness and visual similarity. The results after the attack are shown in Table~\ref{tab.auto}.

\begin{table}
\centering
\caption{Accuracy performance of CNNs on the Flower dataset and the effectiveness of AutoZOOM. Two values are present for the effectiveness of AutoZOOM attack algorithm, one indicating the accuracy of the model after being attacked and the other value in parentheses indicating the fooling rate. (\%)}
\label{tab.auto}
\setlength{\tabcolsep}{11mm}{
\begin{tabular}{lcc} 
\hline
Model & Accuracy & AutoZOOM \\ 
\hline
VGG16 & 90.44 & 73.85\quad(18.34) \\
InceptionV1 & 91.73 & 61.98\quad(32.43) \\
ResNet50 & 88.79 & 72.71\quad(18.11) \\
\hline
\end{tabular}}
\end{table}

Following the same experimental setup, we convert the adversarial examples obtained by AutoZOOM into AFGs. Next, we first verify whether the AFGs generated by the black-box attack have the AFS property used for adversarial example detection. We use 80\% of all AFGs for training and the remaining data for the validation set. The experimental results are shown in Table~\ref{tab.black-det}.

\begin{table}
\centering
\caption{Accuracy of adversarial example detection in AutoZOOM attack. (\%)}
\label{tab.black-det}
\setlength{\tabcolsep}{11mm}{
\begin{tabular}{lcc} 
\hline
AFG detection model & Train Accuracy & Test Accuracy \\ 
\hline
VGG16 & 100.00 & 96.86 \\
\hline
\end{tabular}}
\end{table}

Compared to Table~\ref{tab.defense}, AFGs for detecting adversarial examples have better performance on black-box attacks. This may be due to the fact that the black-box attack does not have access to the model specific parameters, thus making adversarial examples more different from the original images.

We use these generated AFGs to verify whether proposed adversarial example recognition framework can recover the correct labels of adversarial examples under black-box attacks. The experimental results are shown in Table~\ref{tab.black-cla}.

\begin{table}
\centering
\caption{Performance of adversarial example recognition framework under AutoZOOM attack. (\%)}
\label{tab.black-cla}
\setlength{\tabcolsep}{11mm}{
\begin{tabular}{lcc} 
\hline
AFG recognition model & Train Accuracy & Test Accuracy \\ 
\hline
VGG16 & 96.35 & 78.39 \\
\hline
\end{tabular}}
\end{table}

The experimental results demonstrate that the adversarial example recognition framework still correctly classifies adversarial examples. In summary, both Table~\ref{tab.black-cla} and Table~\ref{tab.black-det} show that our proposed recognition framework is still effective for black-box attacks.

\end{document}